\documentclass[10pt,twocolumn,letterpaper]{article}

\usepackage[pagenumbers]{cvpr} 
\usepackage[T1]{fontenc}
\usepackage{amssymb} 
\usepackage{pifont}

\definecolor{cvprblue}{rgb}{0.21,0.49,0.74}
\usepackage[pagebackref,breaklinks,colorlinks,allcolors=cvprblue]{hyperref}

\def\paperID{3440} 
\def\confName{CVPR}
\def\confYear{2025}

\title{Dataset Distillers Are Good Label Denoisers In the Wild}

\author{Lechao Cheng\\
Hefei University of Technology\\
{\tt\small chenglc@hfut.edu.cn}
\and
Kaifeng Chen\\
Zhejiang University of Technology\\
{\tt\small 221123120336@zjut.edu.cn}
\and
Shengeng Tang\\
Hefei University of Technology\\
{\tt\small tangsg@hfut.edu.cn}
\and
Shufei Zhang\\
Shanghai AI Lab\\
{\tt\small zhangshufei@pjlab.org.cn}
\and
Meng Wang\\
Hefei University of Technology\\
{\tt\small wangmeng@hfut.edu.cn}
}

\begin{document}
\maketitle
\begin{abstract}
Learning from noisy data has become essential for adapting deep learning models to real-world applications. Traditional methods often involve first evaluating the noise and then applying strategies such as discarding noisy samples, re-weighting, or re-labeling. However, these methods can fall into a vicious cycle when the initial noise evaluation is inaccurate, leading to suboptimal performance. To address this, we propose a novel approach that leverages dataset distillation for noise removal. This method avoids the feedback loop common in existing techniques and enhances training efficiency, while also providing strong privacy protection through offline processing. We rigorously evaluate three representative dataset distillation methods (DATM, DANCE, and RCIG) under various noise conditions, including symmetric noise, asymmetric noise, and real-world natural noise. Our empirical findings reveal that dataset distillation effectively serves as a denoising tool in random noise scenarios but may struggle with structured asymmetric noise patterns, which can be absorbed into the distilled samples. Additionally, clean but challenging samples, such as those from tail classes in imbalanced datasets, may undergo lossy compression during distillation. Despite these challenges, our results highlight that dataset distillation holds significant promise for robust model training, especially in high-privacy environments where noise is prevalent. The source code is available at \href{https://github.com/Kciiiman/DD_LNL}{https://github.com/Kciiiman/DD\_LNL}.

\end{abstract}    
\section{Introduction}
Learning from noisy data~\cite{natarajan2013learning, zhang2018generalized,li2020dividemix,bae2022noisypredictiontruelabel,englesson2024robust} has emerged as an effective strategy for customizing cutting-edge deep learning algorithms to vertical applications. Existing approaches~\cite{han2018co, shu2019metaweightnetlearningexplicitmapping, yu2018learningbiasedcomplementarylabels, yao2021dualtreducingestimation, ciortan2021frameworkusingcontrastivelearning, wang2024pidualusingprivilegedinformation} tackling this issue generally follows a core step of first assessing the data noise, and then either discarding noisy samples~\cite{han2018co,malach2018decouplingwhenupdatehow,wang2018iterativelearningopensetnoisy}, applying weighted adjustment~\cite{shu2019metaweightnetlearningexplicitmapping,ren2019learningreweightexamplesrobust, zhou2024l2blearningbootstraprobust,bae2024dirichletbasedpersampleweightingtransition}, or reassigning high-confidence labels~\cite{li2023disc, liu2022adaptiveearlylearningcorrectionsegmentation}. Most of the approaches primarily focus on the separation of clean and noisy samples while aiming to prevent overfitting to limited clean data and mitigate biasing model with noisy information.

\begin{figure}
    \centering
    \includegraphics[width=\linewidth]{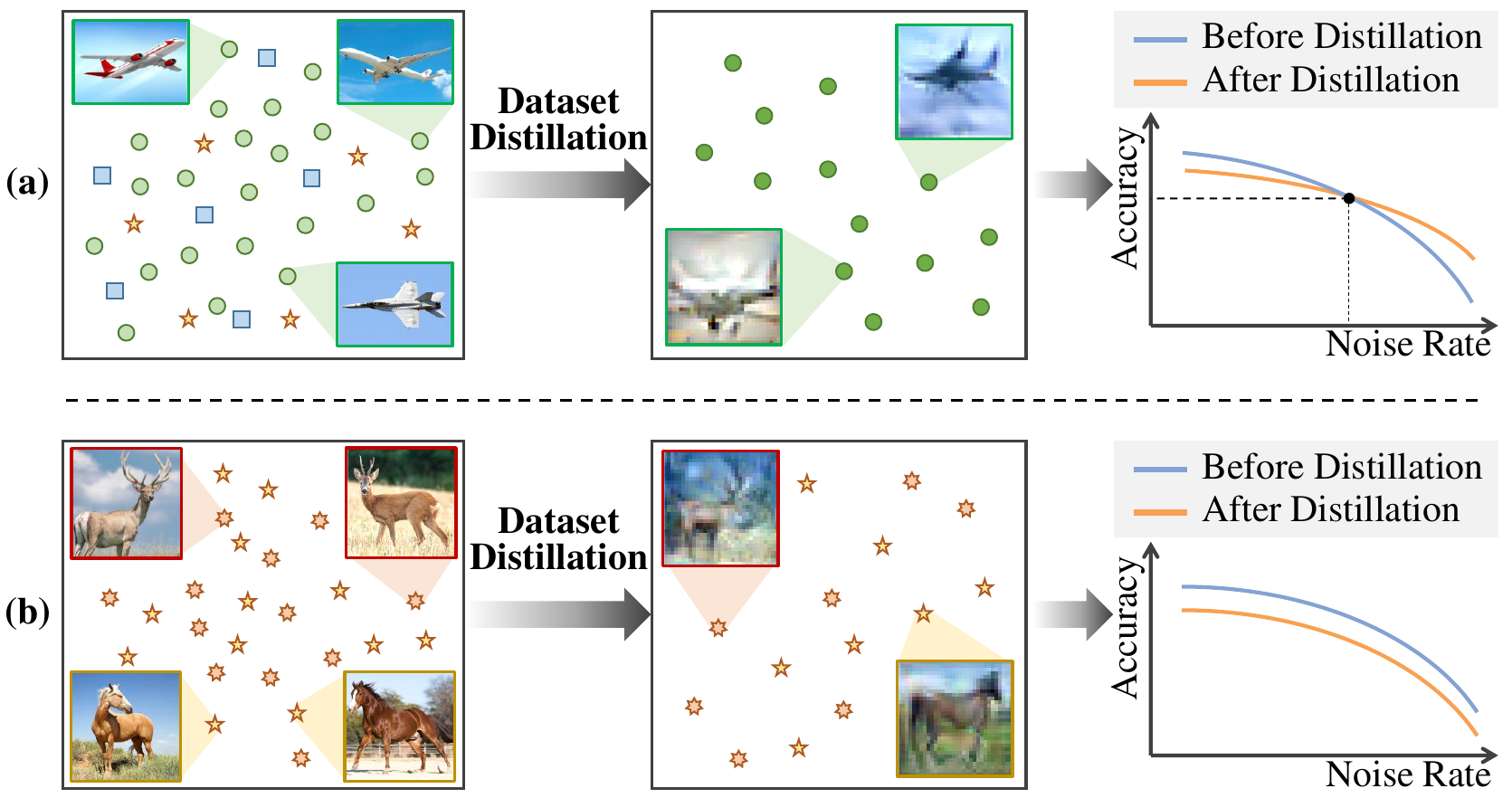}
    \caption{(a) for symmetric noise, existing dataset distillation methods serve as effective denoising tools. (b) structured patterns, such as asymmetric noise, can also be absorbed into the distilled samples, which hinders robust training.}
    \vspace{-5mm}
    \label{fig:teaser}
\end{figure}

\noindent\textbf{Status Quo.} Although numerous existing methods have achieved remarkable results on nearly ideal datasets (such as noisy data curated from CIFAR-10/100~\cite{krizhevsky2009learning}), as mentioned earlier, the general approach involves first evaluating the noise, followed by optimizations such as discarding, re-weighting, or re-labeling. We argue that in such a strategic feedback loop, the noise evaluation process itself is inherently indeterminate in terms of causality: the evaluation of noise affects subsequent strategies (e.g., re-labeling), and those strategies in turn influence the noise evaluation. If the initial noise evaluation is poor, it leads to an inescapable vicious cycle. Moreover, a good noise evaluation source can itself fall into a new paradox, akin to: \textbf{\textit{"It's a Catch-22: You need generalizability to get a good assessment, but you need assessment to gain generalizability"}}.

\noindent\textbf{Motivation.} Given this, let us step outside the existing framework and revisit this task. The core issue here is that we need to use the information from the clean samples to train a robust model while also avoiding bias from noisy data. \textit{Can we, perhaps, sample from data distribution to obtain a support set\footnote{A synthetic subset that can effectively cover the original clean data distribution.} within the manifold of clean sample space ?} This would allow us to directly leverage these support sets to maximize the retention of clean sample information and train a more reliable model\footnote{for now, we will set aside the potential role of slight noise perturbations in promoting robustness}. This raises an intriguing question, and coincidentally, the idea behind dataset distillation~\cite{wang2018dataset} seems to share a similar spirit. Dataset distillation~\cite{wang2018dataset} aims to condense a large dataset into a significantly smaller synthetic dataset that retains the essential information necessary for models to achieve performance comparable to training on the full dataset. In contrast to the "memorization" process~\cite{arpit2017closer, han2018co, yu2019does, li2020gradient, han2020sigua, xia2020robust} in conventional deep learning networks, dataset distillation attempts to synthesize common patterns shared across the dataset, then gradually distills samples with more specific information. This insight has inspired us to explore the use of dataset distillation techniques for noise sample removal.

\noindent\textbf{Empirical Insights.} To further investigate the feasibility of motivation, we first review three representative benchmarking dataset distillation methods, namely parameter matching (DATM~\cite{guo2023towards}), distribution matching (DANCE~\cite{zhang2024dancedualviewdistributionalignment}), and meta-learning (RCIG~\cite{loo2023datasetdistillationconvexifiedimplicit}). These methods are then rigorously validated under commonly used noise scenarios, including symmetric noise, asymmetric noise, and natural noise conditions. We expect that doing so can provide novel perspectives in addressing noisy label learning. Empirically, we find some principled yet intuitive insights to achieve robust training: \textit{(I) for symmetric noise, existing dataset distillation methods serve as effective denoising tools. (II) structured patterns, such as asymmetric noise, can also be absorbed into the distilled samples, which hinders robust training. (III) for real-world natural noise, dataset distillation methods remain effective even in the presence of an unknown fixed noise rate.} Although (I) provides effective guidance for random noise removal, the relationship between the amount of data to be distilled and the level of noise remains an open question worth exploring (Figure~\ref{fig:teaser}(a)). At the same time, (II) reminds us that the common patterns memorized in deep learning are not always beneficial (they may also arise from structured asymmetric noise due to factors like visual similarity), as illustrated in Figure~\ref{fig:teaser}(b). In real-world scenarios, data noise is largely random, and dataset distillation methods can clearly serve as effective denoising techniques. These methods hold significant potential for application in high-privacy settings where data noise is prevalent.

\noindent\textbf{Contributions.} We summarize major contributions bellow:
\begin{itemize}
    \item We propose a new perspective for addressing the problem of model learning from noisy data. This perspective not only effectively avoids the issue of vicious cycles, but also improves training efficiency and provides strong privacy protection through offline dataset distillation processing.
    \item In this new perspective, we conduct in-depth explorations of existing dataset distillation methods and find that, in the context of random noise, dataset distillers serve as effective label denoisers. This conclusion also holds true for natural noise data in a broader sense.
    \item We further investigate potential pitfalls, such as the paradigm of dataset distillation focuses on compressing common patterns, which are not always beneficial. For example, structured asymmetric noise patterns are likely to be distilled into the synthesized data. Additionally, clean yet challenging samples, such as those from tail classes in imbalanced datasets, are at risk of being lossy compressed during the dataset distillation process.
\end{itemize}

\section{Realted Works}

\subsection{Learning with Noisy Labels}
Noisy label learning addresses mislabeled data caused by human error, automated systems, or inherent ambiguity in real-world datasets. Several approaches have been proposed to tackle this challenge~\cite{zhang2018generalized, li2020dividemix, bae2022noisypredictiontruelabel, natarajan2013learning}. Some methods correct noisy labels by modeling the noise transition matrix, which describes label corruption~\cite{yao2021dualtreducingestimation, yu2018learningbiasedcomplementarylabels}. Techniques such as~\cite{cheng2020learning, yang2022estimatinginstancedependentbayeslabeltransition} estimate instance-dependent matrices using deep networks to improve label handling. Additionally,~\citet{wang2024pidualusingprivilegedinformation} leverages privileged information for better noise identification. Contrastive learning improves robustness by contrasting noisy samples with reliable ones, reducing sensitivity to label noise~\cite{ciortan2021frameworkusingcontrastivelearning, ghosh2021contrastivelearningimprovesmodel}. Methods like Twin Contrastive Learning~\cite{huang2023twincontrastivelearningnoisy} build noise-resilient representations, while~\cite{li2022selectivesupervisedcontrastivelearningnoisy} isolates clean data through high-confidence training.
Noise filtering and correction methods, such as dual-network frameworks~\cite{han2018co, malach2018decouplingwhenupdatehow}, iteratively remove noisy samples, while dynamic correction approaches like DISC~\cite{li2023disc} adjust training to prevent overfitting. Reweighting methods~\cite{shu2019metaweightnetlearningexplicitmapping, ren2019learningreweightexamplesrobust} assign adaptive weights to samples based on reliability, reducing the influence of noisy data. Recent methods~\cite{zhou2024l2blearningbootstraprobust, bae2024dirichletbasedpersampleweightingtransition} refine this by incorporating noise modeling and dynamic curricula~\cite{jiang2018mentornetlearningdatadrivencurriculum}, improving robustness.
In particular, approaches~\cite{arpit2017closer, han2018co, yu2019does, li2020gradient, han2020sigua, xia2020robust} such as memorization analysis highlight the tendency of deep networks to overfit noisy data, suggesting the need for methods that reduce the influence of noisy samples. This also motivates us to distill noisy dataset into a compact subset that still enables effective training.


\subsection{Dataset Distillation}
Dataset distillation~\cite{wang2018dataset} seeks to reduce large datasets into smaller synthetic ones that retain the crucial information required for models to perform comparably to those trained on the full dataset. Recent research on dataset distillation has explored several approaches, which can be categorized into meta-learning, parameter matching, and distribution matching~\cite{lei2023comprehensive, geng2023survey, yu2023dataset}. Meta-learning~\cite{hospedales2021meta} involves a two-loop optimization: in the inner loop, the model is trained on the synthetic dataset, and in the outer loop, the dataset is optimized to ensure the model performs well on real data~\cite{wang2018dataset}. Some approaches have enhanced this framework by using soft labels~\cite{bohdal2020flexible, sucholutsky2021soft} or replacing the neural network with kernel models~\cite{loo2022efficient, nguyen2020dataset, nguyen2021dataset}. In contrast, parameter matching methods~\cite{zhao2020dataset, chen2023data, he2024multisize, lee2024selmatch} focus on aligning model parameters trained on both synthetic and real datasets to ensure similar effects. For example,~\citet{zhao2020dataset} use gradient matching during training, while~\citet{he2024multisize} propose a multisize dataset condensation approach based on gradient matching. The trajectory matching approach, exemplified by DATM~\cite{guo2023towards}, aligns training trajectories between synthetic and real datasets, further proposing that the synthetic dataset size should match the difficulty of learning the generated patterns. Additionally, PDD~\cite{chen2023data} incrementally synthesizes groups of synthetic images, training on the union of these subsets, while SelMatch~\cite{lee2024selmatch} addresses large images per class by using selection-based initialization and partial updates through trajectory matching. Distribution matching approaches, such as those by~\citet{zhao2023dataset} using maximum mean discrepancy (MMD), aim to match the feature distributions of synthetic and real datasets, reducing optimization complexity. Improvements to distribution matching~\cite{zhao2023improveddistributionmatchingdataset} incorporate partitioning augmentation and class-aware regularization, enhancing dataset condensation accuracy while potentially reducing diversity. Recent work~\cite{zhang2024dancedualviewdistributionalignment} has focused on modeling inner- and inter-class relationships, with methods like DANCE aligning both feature and label distributions between real and synthetic datasets, further improving model performance. This work first benchmarks existing representative distillation approaches (e.g., DATM~\cite{guo2023towards}, DANCE~\cite{zhang2024dancedualviewdistributionalignment}, RCIG~\cite{loo2023datasetdistillationconvexifiedimplicit}) and then perform exhaustive experiments on noisy data with different ratios. We further present the observations and finally give some insightful conclusions.

\section{Setup and Protocol}
We first present an introduction of the prevailing settings for learning from noisy data, including mainstream noise types, commonly used noisy datasets, and essential information derived from the training protocol.

\subsection{Problem Setup}\label{sec:setup}
\noindent\textbf{Noise Type} This work mainly explores three types of noises, that is \textit{Symmetric Noise}, \textit{Asymmetric Noise}, and \textit{Natural Noise}. Both of which are widely employed in existing researches~\cite{zhang2018generalized, li2020dividemix}. Here we give a brief review.

\begin{itemize}
    \item \textbf{Symmetric Noise}: Labels are randomly altered to any other class $y' \neq y$ with a fixed probability. Formally, 
    \begin{equation}
        p(\tilde{y} = y' | y) = 
        \begin{cases} 
         1 - \tau, & \text{if } y' = y \\
         \frac{\tau}{C - 1}, & \text{if } y' \neq y 
         \end{cases}
    \end{equation}
where $\tilde{y}$ is the noisy label, $C$ is the total number of classes, and $\tau$ represents the noise rate. This noise model simulates uniform random label disturbances, wherein every label has an equal likelihood of being randomly changed to any other class, creating a noise structure that is unbiased across classes.
    \item \textbf{Asymmetric Noise}: This form of noise simulates scenarios where label ambiguity is influenced by inherent class similarities or overlapping boundaries between classes. Mathematically, asymmetric noise can be defined by a conditional probability distribution $p(\tilde{y} = y' \mid y) $ that assigns higher probabilities to semantically similar classes. Formally, let the noise probability $ \tau(y \rightarrow y') $ denote the likelihood of a clean label $y$ being corrupted to a specific label $y'$, where $\tau(y \rightarrow y') $ depends on the semantic similarity between $y$  and $y'$ such that:
    \begin{equation}
        p(\tilde{y} = y' \mid y) = 
            \begin{cases} 
                1 - \sum_{y' \neq y} \tau(y \rightarrow y'), & \text{if } y' = y \\
                \tau(y \rightarrow y'), & \text{if } y' \neq y 
            \end{cases}
    \end{equation}
where $\tilde{y}$ represents the noisy label, and $\tau(y \rightarrow y')$ reflects an asymmetric noise rate specific to pairs of classes  $(y, y')$ . This distribution emphasizes the higher probability of corruption toward semantically close or visually similar classes, capturing realistic noise patterns that occur in scenarios with ambiguous or related class structures.

    \item \textbf{Natural Noise}:Noise in data often originates from the annotation process. Notable datasets that capture natural noise include the CIFAR-N series~\cite{wei2021learning}, which closely approximates real-world label noise by emulating human annotation inconsistencies through multiple settings (Aggregate, Random1-3, and Worst). Additionally, high-volume datasets like Clothing1M~\cite{xiao2015learning} and WebVision~\cite{li2017webvision} serve as classic benchmarks for studying learning under realistic label noise conditions, providing extensive resources for robust learning research.
\end{itemize}

\noindent\textbf{Training Protocol}
Noisy label learning aims to effectively learn from data where labels are often corrupted or inaccurate. Let $\mathcal{\tilde{S}}=\{x_i, \tilde{y_i}\}_{i=1}^{n}$ denote a dataset with noisy labels, where $x_i$ is the input image, and $\tilde{y_i}$ indicates the potentially corrupted label. The dataset $\mathcal{\tilde{S}}$ is sampled from a noisy joint distribution $\mathcal{\tilde{D}}$ distinct from the clean one $\mathcal{D}$. For this task, Cross-Entropy loss (CE) is commonly used as the loss function to train a robust classifier $f: X \rightarrow Y$ capable of assigning the true label $y$ to test instances. Formally, 
\begin{equation}
    \mathcal{L}_{CE}(f(x), y) = -\sum_{c=1}^{C} y_c log(f_c(x)),
\end{equation}
where $C$ is the total number of classes, $y_c$ is the label for class $c$, and $f_c(x)$ indicates the model predicted probability for class $c$.

\noindent\textbf{Remark.} Existing work~\cite{arpit2017closer} has explored the effect of "memorization" in DNNs, especially in noisy label learning~\cite{han2018co, yu2019does, li2020gradient, han2020sigua, xia2020robust}. This kind of phenomenon reveals that networks always first memorize training data of common patterns (e.g. clean labels) and then those of hard samples (e.g., noisy labels). In this work, we propose a novel perspective that the process of dataset distillation aims to condense informatively subset data that shares the similar spirit of learning from noisy labels, both of which attempt to promote stable model training with common patterns. The distinction lies in their objectives: noise learning aims at achieving a robust learning model, whereas dataset distillation seeks to capture reliable, clean, and common patterns.
This raises an important question: \textbf{\textit{ Why not directly distill a clean subset that enables effective training?}} This is intuitive, and we begin benchmarking existing methods and provide insights based on experimental results in the following section.

\subsection{Benchmarking Dataset Distillation}
We observe a recent surge in research on dataset distillation~\cite{wang2018dataset}, making it challenging to cover all related works comprehensively. Thus, we select three representative state-of-the-art methods based on their categorization: distribution matching (DANCE~\cite{zhang2024dancedualviewdistributionalignment}), meta-learning (RCIG~\cite{loo2023datasetdistillationconvexifiedimplicit}), and parameter matching (DATM~\cite{guo2023towards}).

\noindent\textbf{Benchmark-I: DATM} is an advanced representative parameter matching method grouped on trajectory matching. It optimizes the generation of synthetic datasets by directly matching the model parameters trained on synthetic datasets with those trained on real datasets. In this way, DATM assumes that the performance of synthetic dataset can be achieved by aligning with the training trajectory of the real dataset, thus improving the quality of synthetic data. Specifically, let us define $\theta_t^*$ as the expert model parameters obtained after $t$ steps of training on real images, and $\theta_{t+M}$ denotes the expert model parameters after $M$ steps after $\theta_t^*$. Similarly, $\hat{\theta}_{t+T}$ denotes the student model parameters obtained after $T$ steps of training on the synthetic dataset, initialized from $\theta_t^*$. The objective of DATM is to align the student model trained for $T$ steps on the synthetic data with the performance of the expert model trained for $M$ steps on real data, typically under the condition $T << M$. Formally, this objective is defined as follows:
\begin{equation}
    \mathcal{L} = \frac{\| \hat{\theta}_{t+N} - \theta_{t+M}^* \|_2^2}{\| \theta_t^* - \theta_{t+M}^* \|_2^2}.
\end{equation}
While DATM demonstrates strong performance on smaller datasets, it faces significant scalability challenges on large-scale, real-world datasets like ImageNet. Recent work proposed by~\citet{cui2023scaling} identifies the bottleneck in trajectory matching as the unrolled gradient computation, revealing that the overall memory complexity remains constant with respect to $T$. The proposed memory-efficient variant requires only a single gradient computational graph, thereby drastically reducing computational overhead while maintaining comparable performance to the original method. Unless otherwise specified, \textbf{Benchmark-I} referenced later denotes this accelerated version.

\noindent\textbf{Benchmark-II: DANCE} improves the milestone distribution matching method by addressing both the inner-class and the inter-class limitations. Formally, the condensed set is optimized by:
\begin{equation}
\mathcal{D}_{\text{syn}}^* = \arg \min_{\mathbb{E}_{\phi} \sim P_{\phi}} \| \frac{\sum_{i=1}^{|\mathcal{D}_{\text{real}}|}\phi(\text{x}_i^{\text{real}})}{\mathcal{D}_{\text{real}}}- \frac{\sum_{j=1}^{|\mathcal{D}_{\text{syn}}|}\phi(\text{x}_j^{\text{syn}})}{\mathcal{D}_{\text{syn}}}\|^2, 
\end{equation}
where $\phi$ combines the randomly initialized encoders $\phi_0$ and their corresponding trained counterpart $\phi_{\text{expert}}$, that is,
\begin{equation}
    \phi \leftarrow \lambda.\phi_0 + ( 1 - \lambda).\phi_{\text{expert}}.
\end{equation}
Herein $\lambda \sim U(0,1)$ is a randomly generated value. Apart from this, \textbf{DANCE} also pushes parameters of $\phi_{\text{expert}}$ to adapt synthetic data to calibrate the inter-class distribution shift:
\begin{equation}
    \mathcal{L}_{\text{expert}} = \frac{1}{\mathcal{D}_{\text{syn}}}\sum_{j=1}^{|\mathcal{D}_{\text{syn}}|} \ell_{ce}(\theta_{\text{expert}}(x_j^{\text{syn}}),y_j^{\text{syn}}).
\end{equation}
This kind of strategy regularizes the synthetic data near the original data distribution. More details can be found in~\cite{zhang2024dancedualviewdistributionalignment}.

\begin{figure*}[!htb]
    \centering
    \begin{subfigure}[b]{0.9\textwidth}
        \centering
        \includegraphics[width=\textwidth]{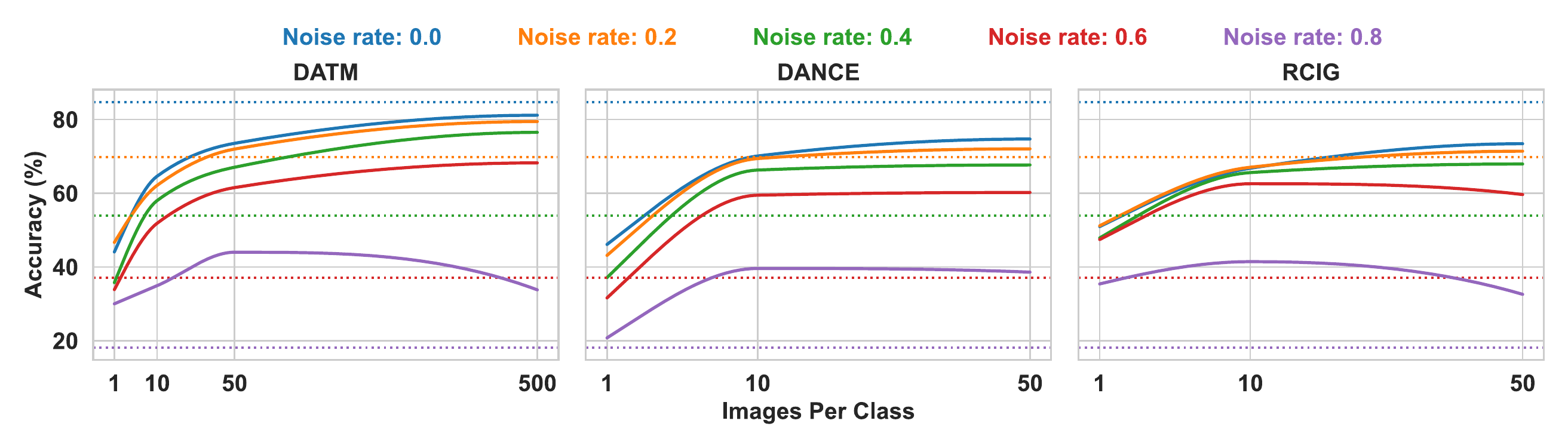}
        \caption{Symmetric noisy dataset distillation for different noisy ratio on \textbf{CIFAR-10}. }
        \label{fig:cifar10_ipc_sym_last}
    \end{subfigure}

    \begin{subfigure}[b]{0.9\textwidth}
        \centering
        \includegraphics[width=\textwidth]{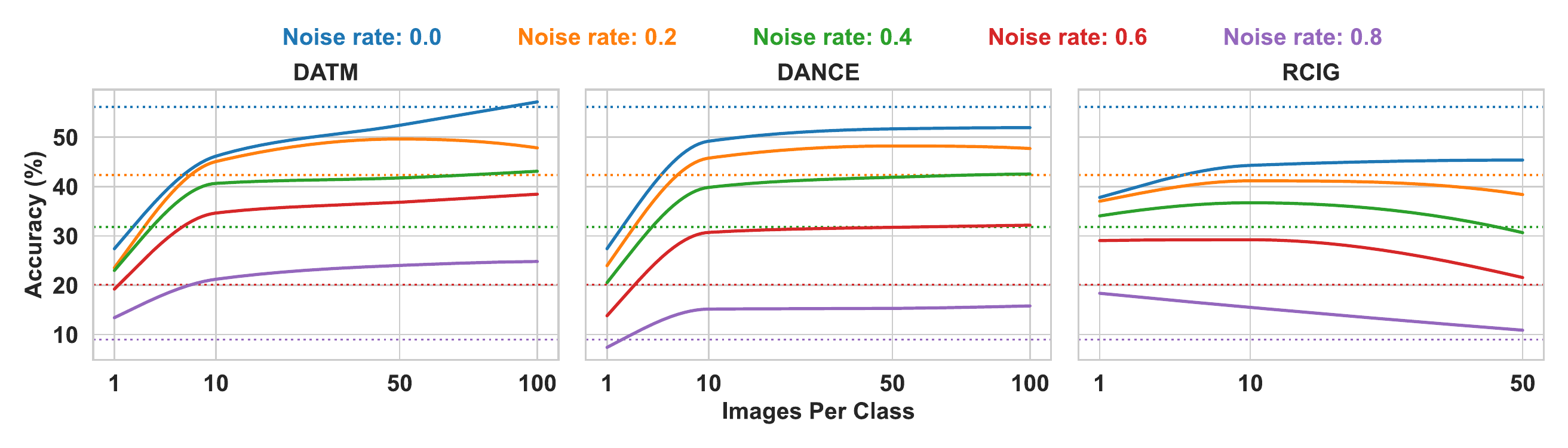}
        \caption{Symmetric noisy dataset distillation for different noisy ratio on \textbf{CIFAR-100}.}
        \label{fig:cifar100_ipc_sym_last}
    \end{subfigure}

    \begin{subfigure}[b]{0.9\textwidth}
        \centering
        \includegraphics[width=\textwidth]{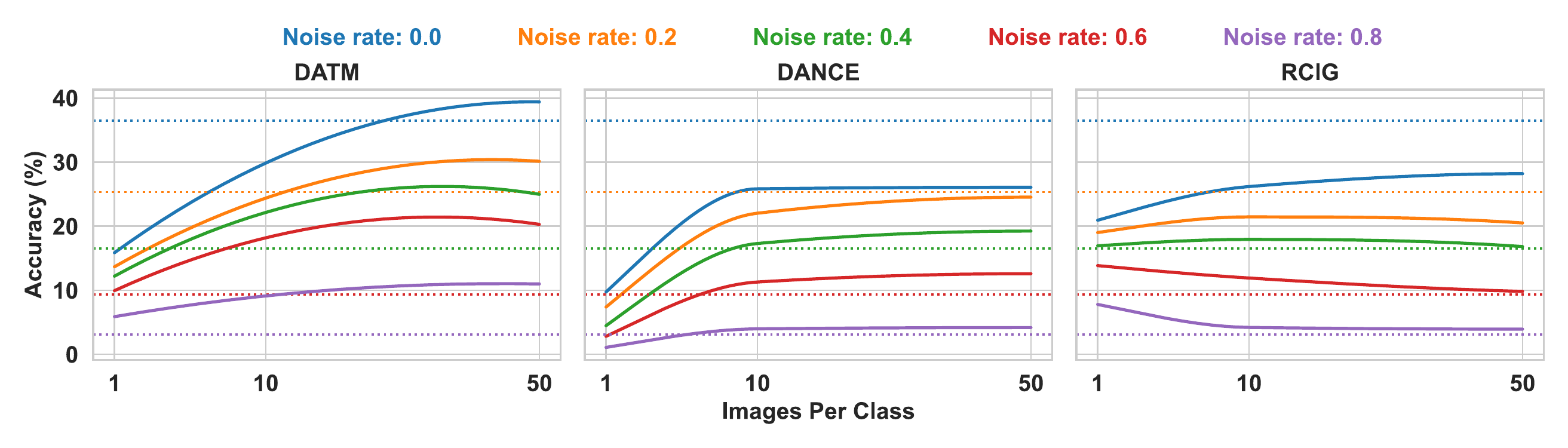}
        \caption{Symmetric noisy dataset distillation for different noisy ratio on \textbf{Tiny-ImageNet}.}
        \label{fig:tiny_ipc_last}
    \end{subfigure}
    
    \caption{The validation performance over symmetric noise for CIFAR-10, CIFAR-100, and Tiny-ImageNet datasets.
    The solid lines represent the accuracy trend as Image Per Class (IPC) increases, while the horizontal dashed lines of the same color indicate the performance of training on the full dataset at the corresponding noise rate.
    }
    \label{fig:sym_ipc_comparison}
\end{figure*}

\noindent\textbf{Benchmark-III: RCIG }proposes a novel meta-learning framework for dataset distillation. It is achieved by reparameterizing and convexifying implicit gradients to enable analytical exploration. Let $\mathcal{L}_o = \mathcal{L}_T(\theta), \mathcal{L}_i = \mathcal{L}_{S(\psi)}(\theta)$. $\mathcal{L}_T(\theta)$ and $\mathcal{L}_{S(\psi)}(\theta)$ denote the training losses of the full training set and the support set, respectively. $\psi$ is just the goal of dataset distillation.  Specifically, \textbf{RCIG} as a meta-learning method, defines its inner and outer loops as follows:
\begin{equation}
\begin{split}
        & \mathcal{L}_i(\theta_B) = \mathcal{L}_{S(\psi)}(\theta_B, \theta_F^*(\theta_B, \psi)) \\
        &\mathcal{L}_o(\theta_B,\psi) = \mathcal{L}_{\text{platt},T}(\theta_B, \theta_F^*(\theta_B, \psi),\tau).
\end{split}
\end{equation}
Here, $\theta_B$ and $\theta_F$ denote the backbone parameters and the final layer parameters, respectively. $\tau$ is the learnable temperature parameter. $\mathcal{L}_{\text{platt}}$ originates from Platt scaling loss~\cite{platt1999probabilistic}. $\theta_F^*$ is defined with neural network gaussian process as:
\begin{equation}
\begin{split}
    \theta_F^* &= h_{\theta_0}(X_S)(K_{X_S,X_S}^{\theta_0} + \lambda I_{|S|})^{-1}\hat{y}_S \\
    \hat{y}_S &= (y_S - \theta_B^T\frac{\partial f_{\text{lin},\theta}(X_S)}{\partial \theta_B}),
\end{split}
\end{equation}
where $y_S$ means labels and $f_{\text{lin}}$ is the 1st-order Taylor approximation of learning dynamics linearized dynamics~\cite{loo2022evolution}. $h_{\theta_0}(X_S)$ defines the embeddings of the hidden layer, with $|S|$ being the distilled data size and $H$ the final layer dimension. Noted that more information about NTK $K_{X_S,X_S}^{\theta_0}$ can be found in~\cite{loo2022evolution, zhou2022dataset}.

\noindent\textbf{Remark.} In this section, we revisit three types of dataset distillation benchmarking approaches. We further explore these representative methods to validate whether the distillation procedure works well on noisy data. We believe that the results of these experiments can provide insightful perspectives for offering an innovative solution to the data noise challenge but also demonstrate substantial potential for application in privacy preservation. 

\section{Observations and Insights}
\begin{figure*}[!htb]
    \centering
    \includegraphics[width=0.9\textwidth]{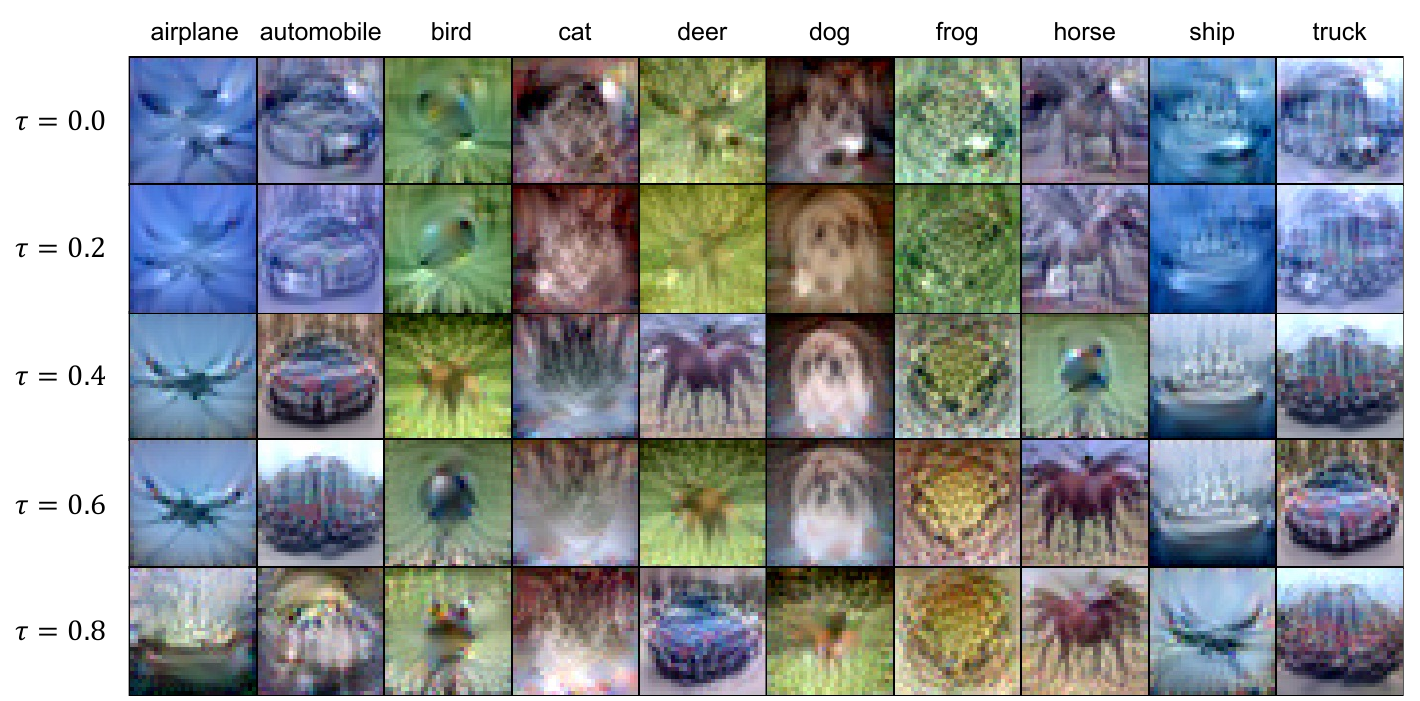}
    \caption{Visualization of images distilled from DATM on CIFAR-10 with one image per class.}
    \label{vis:cifar10_noise_last}
\end{figure*}
This section follows a structured pipeline: we begin by outlining the experimental details, followed by a summary of key observations, and conclude with insightful conclusions.

\noindent \textbf{Implementations.} We evaluate these benchmarks using datasets CIFAR-10/100~\cite{krizhevsky2009learning}, and tiny-ImageNet~\cite{le2015tiny}, curating noisy versions~\cite{patrini2017makingdeepneuralnetworks,zhang2018generalized} using symmetric and asymmetric described in Sec.~\ref{sec:setup}. Specifically, for asymmetric noise, labels are flipped to similar classes (e.g., in CIFAR-10: TRUCK $\rightarrow$ AUTOMOBILE, BIRD $\rightarrow$ AIRPLANE, DEER $\rightarrow$ HORSE, CAT $\longleftrightarrow$ DOG; in CIFAR-100, the 100 classes are grouped into 20 superclasses, with each subclass flipping to the next within the same superclass). Additionally, we also adopt a more challenging version CIFAR-N~\cite{wei2021learning} that mimics human annotations.
Following~\cite{zhao2021datasetcondensationdifferentiablesiamese, zhao2023dataset, cazenavette2022datasetdistillationmatchingtraining}, we employ a simple ConvNet~\cite{sagun2018empiricalanalysishessianoverparametrized} architecture for distillation: a three-layer ConvNet for CIFAR and a four-layer ConvNet for Tiny-ImageNet. Performance is evaluated based on test accuracy on distilled datasets, following the evaluation protocols of DATM, DANCE, and RCIG, with data augmentation applied for RCIG as recommended in the original work. Final test accuracies are reported throughout the distillation process.

\noindent  \textbf{Experiment-I: }\ding{43} \textbf{Symmetric Noise Distillation}

Figure~\ref{fig:sym_ipc_comparison} shows performance across different IPC (Images Per Class) and noise conditions under symmetric noise for CIFAR-10, CIFAR-100, and Tiny-ImageNet. We use Cross-Entropy loss by default to investigate the impact of dataset distillation on noisy data. The dashed lines in each subplot represent evaluations of training on the full dataset at the current noise rate.

\noindent \ding{93} \textbf{Observation-I}: \textit{Once the noise surpasses a certain threshold, the three representative dataset distillation methods consistently demonstrate significant performance gains over the baseline trained on the entire noisy dataset, even when distilled to very few samples.}

As shown in Figure~\ref{fig:sym_ipc_comparison}(a), training results after dataset distillation consistently outperform the baseline at noise rates $\tau = {0.2, 0.4, 0.6, 0.8}$. As noise increases, distillation achieves better performance with fewer distilled samples. For example, at noise rates of 0.6 and 0.8, a single distilled sample per class surpasses the baseline. At a noise rate of 0.2, fewer than 50 samples per class significantly outperform the baseline (indicated by the yellow curve and dashed line). Similar trends are observed in Figures~\ref{fig:sym_ipc_comparison}(b) and (c). Figure~\ref{vis:cifar10_noise_last} shows the results of distilled DATM images on CIFAR-10, where a single distilled image retains discriminative features despite high noise ratios.
\begin{figure*}[!htb]
    \centering
    \begin{subfigure}[b]{0.9\textwidth}
        \centering
        \includegraphics[width=\textwidth]{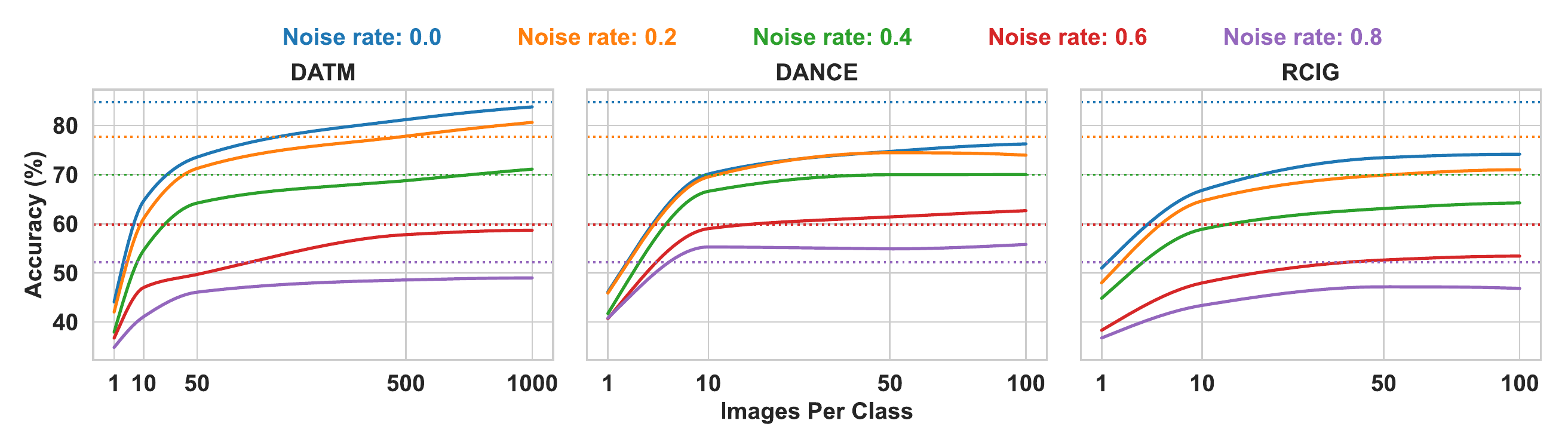}
        \caption{Asymmetric noisy dataset distillation for different noisy ratios on \textbf{CIFAR-10}. }
        \label{fig:cifar10_asy_ipc_last}
    \end{subfigure}

    \begin{subfigure}[b]{0.9\textwidth}
        \centering
        \includegraphics[width=\textwidth]{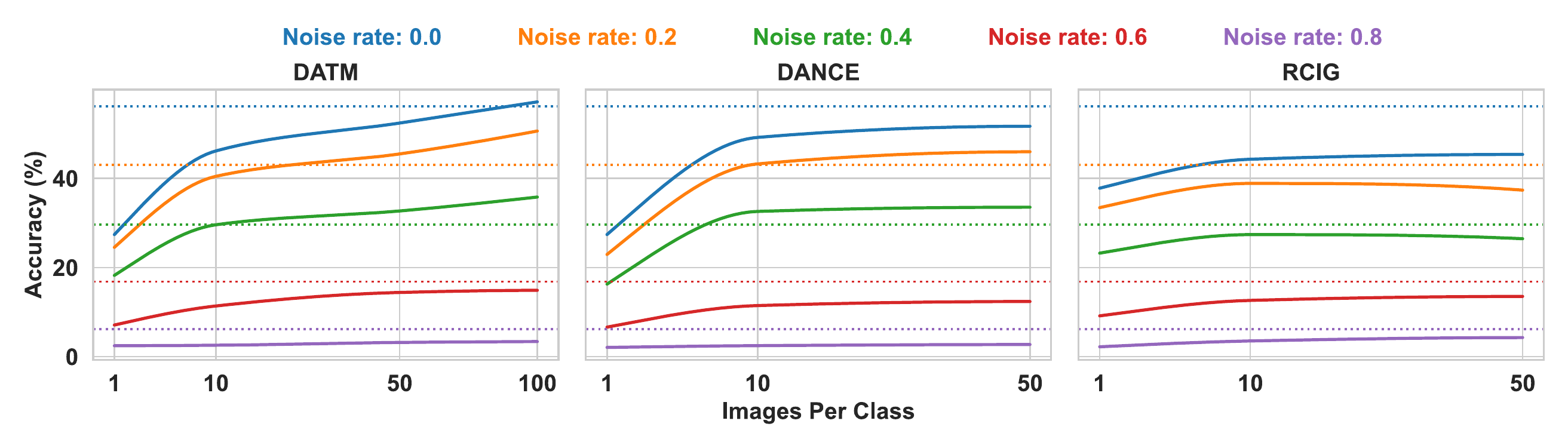}
        \caption{Asymmetric noisy dataset distillation for different noisy ratios on \textbf{CIFAR-100}.}
        \label{fig:cifar100_asy_ipc_last}
    \end{subfigure}
    
    \caption{The validation performance over asymmetric noise for CIFAR-10, CIFAR-100.
    The solid lines represent the accuracy trend as Image Per Class (IPC) increases, while the horizontal dashed lines of the same color indicate the performance of training on the full dataset at the corresponding noise rate}
    \label{fig:cifar_asym_ipc_comparison}
\end{figure*}

\noindent \ding{45} \textbf{Insight-I}: For symmetric noise, existing dataset distillation methods serve as effective denoising tools, which supports with our initial intuitive assumption: \textbf{the distillation process primarily captures common patterns while ignoring outliers such as noise}. However, the critical question remains: How much data should be distilled from a noisy dataset to create a meaningful distilled set? This question could potentially be answered by analyzing the results of dataset distillation, offering a way to estimate the proportion of noise in the data. This presents a compelling and valuable problem for further analysis and exploration.

\noindent \textbf{Corollary-I.} \textit{Given a balanced dataset $\mathcal{\tilde{S}} \in \mathcal{\tilde{D}}$ with symmetric noise $\tau$, to perfectly preserve the sample information from the original dataset for each class during dataset distillation, the maximum number of distilled samples per class (Image Per Class, IPC) required is:}
\begin{equation}
    \text{IPC} \leq \frac{|\mathcal{\tilde{S}}| \cdot (1 - \tau)}{C},
\end{equation}
\noindent\textit{where $C$ is the total number of classes. This corollary succinctly states the upper bound for the number of distilled samples per class needed to retain clean data information under the given noise conditions.}

\noindent \textbf{Experiment-II: }\ding{43} \textbf{Asymmetric Noise Distillation}

We also perform dataset distillation experiments on CIFAR-10 and CIFAR-100 with asymmetric noise. As illustrated in Figure~\ref{fig:cifar_asym_ipc_comparison}, most state-of-the-art methods perform worse on datasets with asymmetric noise compared to the benchmark results from training on the full-noise dataset. Only the \textbf{DATM} method achieves comparable performance to the benchmark at noise rate \(\tau = \{0.2, 0.4\}\), provided the distilled synthetic dataset is sufficiently large.

\noindent \ding{93} \textbf{Observation-II}: \textit{Dataset distillation methods struggle to synthesize the samples that cover the original clean data distribution when confronted with asymmetric noise. Even when a larger number of synthetic samples are distilled, they still fail to accurately capture the true clean data distribution.}

This phenomenon is consistent with intuition. Specifically, in our data construction process, $\tau(y \rightarrow y')$ denotes the probability of corrupting the label of a clean sample $y$ to a noisy label $y'$. Generally, label corruption tends to occur primarily among visually similar classes, such as TRUCK $\rightarrow$ AUTOMOBILE, as discussed before. 
In such cases, the noise process becomes structured or 'patternized'. Consequently, during dataset distillation, this structured noise is inevitably carried over alongside the clean sample patterns, causing the distilled dataset to retain the same label transition patterns. Thus, while the distillation process preserves the clean data patterns, it also captures the inherent noise structure, which hinders the synthetic dataset from accurately representing the true data distribution. 


\begin{figure*}[!htb]
    \centering
    \begin{subfigure}[b]{0.9\textwidth}
        \centering
        \includegraphics[width=\textwidth]{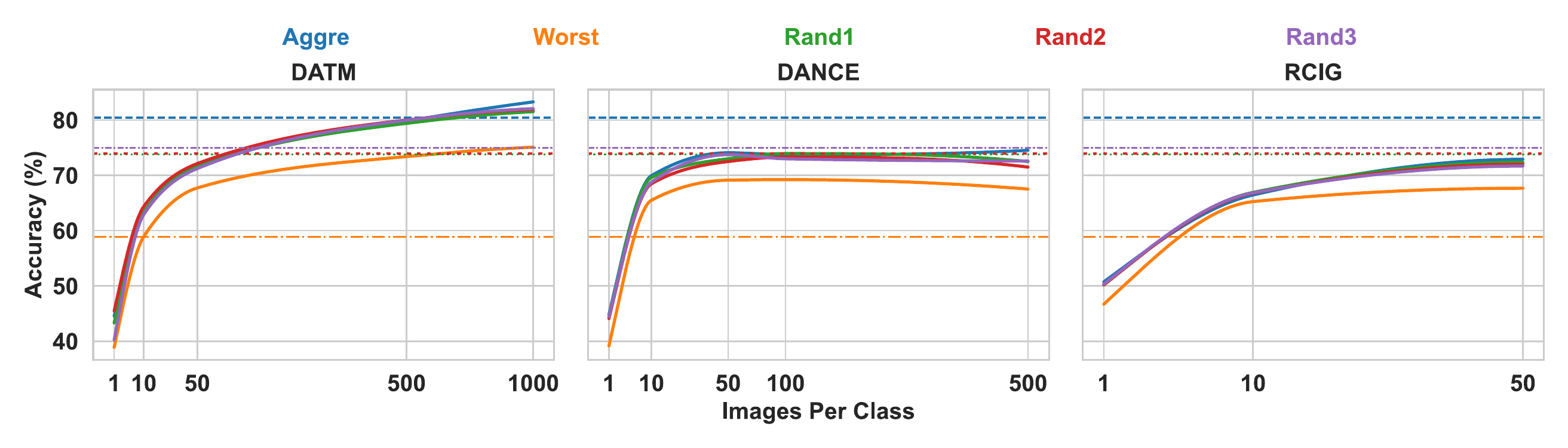}
        \caption{Natural noisy dataset distillation on \textbf{CIFAR-10N}. }
        \label{fig:cifar10n_ipc_last}
    \end{subfigure}

    \begin{subfigure}[b]{0.9\textwidth}
        \centering
        \includegraphics[width=\textwidth]{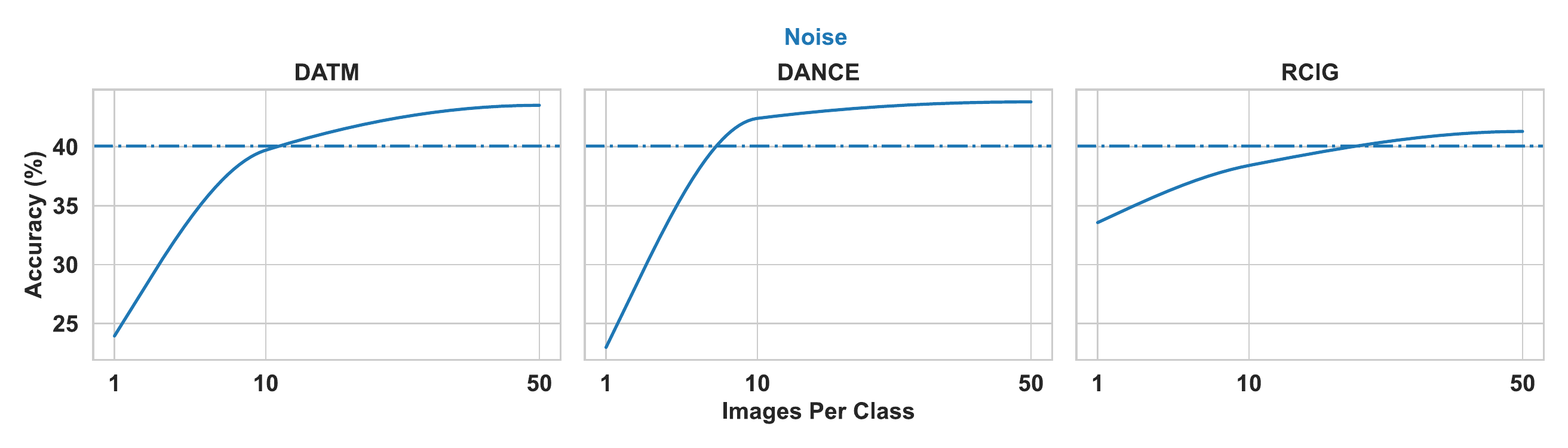}
        \caption{Natural noisy dataset distillation on \textbf{CIFAR-100N}.}
        \label{fig:cifar100n_ipc_last}
    \end{subfigure}
    
    \caption{The results for CIFAR-10N/CIFAR-100N. The solid lines illustrate the trend of accuracy as Image Per-Class (IPC) increases. The horizontal dashed lines of the same color indicate the evaluation of training on the full dataset under the corresponding noise rate.}
    \label{fig:cifarn_ipc_comparison}
\end{figure*}

\noindent \ding{45} \textbf{Insight-II}: This observation provides a valuable insight that warrants deeper exploration. Specifically, when employing dataset distillation to address label noise, it is crucial not to assume that common patterns always reflect clean data. Under certain conditions (e.g., when categories with high visual similarity are mislabelled due to human annotation errors), the resulting noise may be structured rather than purely stochastic. 
This highlights the necessity for a more nuanced approach to dataset distillation, as they may inadvertently preserve and amplify these noise patterns. We further hypothesize that challenging clean samples, such as tail data in imbalanced datasets, are at risk of being lossy compressed during the dataset distillation process.
Given space constraints, we encourage further research into how structured noise/hard clean samples can be identified and well processed during distillation, ensuring that such biases do not undermine the quality of the distilled dataset.

\noindent  \textbf{Experiment-III: }\ding{43} \textbf{Natural Noise Distillation}

Previous experiments have investigated dataset distillation for synthetic noisy labels on CIFAR10, CIFAR100, and Tiny-ImageNet, which are widely used benchmarks. In this part, we empirically validate the robustness on human annotated labels~\cite{wei2021learning} as illustrated in Figure~\ref{fig:cifarn_ipc_comparison}. For CIFAR10N, each training image contains one clean label and three human annotated labels. \textbf{Random k} ($\tau \approx 18\%$) means the $k-$th annotated labels while \textbf{Worst} ($\tau=40.21\%$) and \textbf{Aggre} ($\tau=9.03\%$) indicate the selected wrong label and majority voting label, respectively. For CIFAR100N, we chose the "fine" version from~\cite{wei2021learning} that contains 40.20\% noises.


\noindent \ding{93} \textbf{Observation-III}: \textit{Dataset distillation approaches still generalize well on real-world defective data with fixed noisy ratio, especially for high noise ratio.}

We notice that in Figure~\ref{fig:cifarn_ipc_comparison}, all distillation methods perform well under the \textbf{Worst} scenario, with \textbf{DATM}, \textbf{DANCE}, and \textbf{RCIG} surpassing the CIFAR100N baseline using fewer than 10 distilled images per class. 
However, for lower noise ratio like $\textbf{Aggre}$, these distillation approaches struggle to perform well. 

\noindent \ding{45} \textbf{Insight-III}: For real-world natural noise, dataset distillation methods remain largely effective even in the presence of an unknown fixed noise rate. The challenge in this scenario is similar to that with symmetric noise, specifically the need to determine the appropriate amount of distilled data, i.e., how much valid information should be preserved. When the number of retained samples is small, not only is noisy data excluded, but even good clean samples are subjected to lossy compression. This is why data distillation often yields suboptimal results in low-noise settings.

\noindent \textbf{Corollary-II.} \textit{Given a noisy dataset $\mathcal{\tilde{S}} \in \mathcal{\tilde{D}}$ from a real-world scenario with $C$ classes, when applying dataset distillation such that each class is synthesized to a size of IPC, and the validation accuracy matches that of the original dataset, we can infer with at least high probability $1 - \delta$ that the noise rate $\tau$ of the dataset satisfies:}
\begin{equation}
  \underset{\mathcal{\tilde{S}} \in \mathcal{\tilde{D}}}{\mathbb{P}}\left[\tau \geq (1 - \frac{C \cdot \text{IPC}}{|\mathcal{\tilde{S}}|})\right] \geq 1 - \delta.
\end{equation}
This corollary provides a probabilistic bound on the noise rate $\tau$ in a noisy dataset based on the distillation process and the accuracy achieved after distillation.

\section{Conclusion}
We propose a new approach to model learning from noisy data that avoids vicious cycles, improves training efficiency, and ensures privacy protection through offline dataset distillation. Our findings show that dataset distillation effectively denoises random and natural noise, though it may struggle with structured asymmetric noise and lossy compression of challenging clean samples, especially in imbalanced datasets. Despite these limitations, dataset distillation shows strong potential for robust model training, particularly in high-privacy environments.
{
    \small
    \bibliographystyle{ieeenat_fullname}
    \bibliography{main}
}


\end{document}